\documentclass[11pt,a4paper]{article}
\usepackage{authblk}
\usepackage[hyperref]{mrp19}
\usepackage[T1]{fontenc}
\usepackage{times}
\usepackage{latexsym}
\usepackage{url}
\usepackage{adjustbox}
\usepackage{multirow}
\usepackage{tikz}
\usepackage{graphics}
\usepackage[pdf]{graphviz}

\usetikzlibrary{shapes}

\aclfinalcopy 

\title{HUJI-KU at MRP~2020:\ Two Transition-based Neural Parsers}

\author[*]{\textbf{Ofir Arviv}}
\affil[*]{Hebrew University of Jerusalem, School of Computer Science and Engineering}
\author[**]{\textbf{Ruixiang Cui}}
\author[**]{\textbf{Daniel Hershcovich}}
\affil[**]{University of Copenhagen, Department of Computer Science}
\affil[ ]{\texttt{ofir.arviv@mail.huji.ac.il},\quad\texttt{\{rc,dh\}@di.ku.dk}}

\date{}

\begin{document}
\maketitle
\begin{abstract}
  This paper describes the HUJI-KU system submission to the shared task
  on Cross-Framework Meaning Representation Parsing (MRP) at the 2020
  Conference for Computational Language Learning (CoNLL),
  employing TUPA and the HIT-SCIR parser, which were, respectively,
  the baseline system and winning system in the 2019 MRP shared task.
  Both are transition-based parsers using BERT contextualized embeddings.
  We generalized TUPA to support the newly-added MRP frameworks and languages,
  and experimented with multitask learning with the HIT-SCIR parser.
  We reached 4th place in both the cross-framework and cross-lingual tracks.
\end{abstract}

\section{Introduction}\label{sec:intro}

The CoNLL 2020 MRP Shared Task \cite{Oep:Abe:Abz:20}
combines five frameworks for graph-based meaning representation:
EDS, PTG, UCCA, AMR and DRG.
It further includes evaluations in English, Czech, German and Chinese.
While EDS, UCCA and AMR participated in the 2019 MRP shared task
\cite{Oep:Abe:Haj:19}, which focused only on English,
PTG and DRG are newly-added frameworks to the MRP uniform format.

For this shared task, we extended TUPA \cite{hershcovich2017a},
which was adapted as the baseline system in the 2019 MRP shared task
\cite{hershcovich-arviv-2019-tupa},
to support the two new frameworks and the different languages. In order to add this support, only minimal changes were needed, demonstrating TUPA's strength in parsing a wide array of representations. 
TUPA is a general transition-based parser
for directed acyclic graphs (DAGs),
originally designed for parsing UCCA
\cite{abend2013universal}.
It was previously used as the baseline system in
SemEval 2019 Task 1 \cite{hershcovich2019semeval},
and generalized to support other frameworks \cite{Her:Abe:Rap:18,hershcovich2018universal}.

We also experimented with the HIT-SCIR parser
\cite{che-etal-2019-hit}.
This was the parser with the highest average score across
frameworks in the 2019 MRP shared task, and has also since
been applied to other frameworks 
\cite{hershcovich-etal-2020-kopsala}.

\begin{figure*}[ht]
	\begin{adjustbox}{width=.99\textwidth,margin=1pt,frame}
	\begin{tabular}{llll|l|lllll}
		\multicolumn{4}{c|}{\textbf{\small Before Transition}} &
		\multirow{2}{*}{\textbf{\small Transition}} & 
		\multicolumn{5}{c}{\textbf{\small After Transition}} \\
		\textbf{\footnotesize Stack} & \textbf{\footnotesize Buffer} & 
		\textbf{\footnotesize N.} & \textbf{\footnotesize Edges} & & 
		\textbf{\footnotesize Stack} & \textbf{\footnotesize Buffer} & 
		\textbf{\footnotesize Nodes} & \textbf{\footnotesize Edges} & 
		\textbf{\footnotesize Extra Effect} \\ \hline
		$S$ & $x\;|\;B$ & $V$ & $E$ & \textsc{Shift} & $S\;|\;x$ & $B$ & $V$ & $E$ & \\
		$S\;|\;x$ & $B$ & $V$ & $E$ & \textsc{Reduce} & $S$ & $B$ & $V$ & $E$ & \\
		$S\;|\;x$ & $B$ & $V$ & $E$ & \textsc{Node$_X$} & $S\;|\;x$ & $y\;|\;B$ & $V\cup\{y\}$ & $E\;|\;(y,x)$ & $\ell_E(y,x)\leftarrow X$ \\
		$S\;|\;x$ & $B$ & $V$ & $E$ & \textsc{Child$_X$} & $S\;|\;x$ & $y\;|\;B$ & $V\cup\{y\}$ & $E\;|\;(x,y)$ & $\ell_E(x,y)\leftarrow X$ \\
		$S\;|\;x$ & $B$ & $V$ & $E$ & \textsc{Label$_X$} & $S\;|\;x$ & $B$ & $V$ & $E$ & $\ell_V(x)\leftarrow X$ \\
		$S\;|\;x$ & $B$ & $V$ & $E$ & \textsc{Property$_X$} & $S\;|\;x$ & $B$ & $V$ & $E$ & $p(x)\leftarrow p(x)\cup\{X\}$ \\
		$S\;|\;y,x$ & $B$ & $V$ & $E$ & \textsc{Left-Edge$_X$} & $S\;|\;y,x$ & $B$ & $V$ & $E\;|\;(x,y)$ & $\ell_E(x,y)\leftarrow X$ \\
		$S\;|\;x,y$ & $B$ & $V$ & $E$ & \textsc{Right-Edge$_X$} & $S\;|\;x,y$ & $B$ & $V$ & $E\;|\;(x,y)$ & $\ell_E(x,y)\leftarrow X$ \\
		$S$ & $B$ & $V$ & $E\;|\;(x,y)$ & \textsc{Attribute$_X$} & $S$ & $B$ & $V$ & $E\;|\;(x,y)$ & $a(x,y)\leftarrow a(x,y)\cup\{X\}$ \\
		$S\;|\;x,y$ & $B$ & $V$ & $E$ & \textsc{Swap} & $S\;|\;y$ & $x\;|\;B$ & $V$ & $E$ & \\
		$[\mathrm{root}]$ & $\emptyset$ & $V$ & $E$ & \textsc{Finish} & $\emptyset$ & $\emptyset$ & $V$ & $E$ & terminal state \\
	\end{tabular}
	\end{adjustbox}
	\caption{\label{fig:transitions}
	  The TUPA-MRP transition set, from \newcite{hershcovich-arviv-2019-tupa}.
	  We write the stack with its top to the right and the buffer with its head to the left;
	  the set of edges is also ordered with the latest edge on the right.
	  \textsc{Node}, \textsc{Label}, \textsc{Property} and \textsc{Attribute}
	  require that $x\neq\mathrm{root}$;
	  \textsc{Child}, \textsc{Label}, \textsc{Property},
	  \textsc{Left-Edge} and \textsc{Right-Edge}
	  require that $x\not\in w_{1:n}$;
	  \textsc{Attribute} requires that $y\not\in w_{1:n}$;
	  \textsc{Left-Edge} and \textsc{Right-Edge}
	  require that $y\neq\mathrm{root}$ and
	  that there is no directed path from $y$ to $x$;
	  and \textsc{Swap} requires that $\mathrm{i}(x)<\mathrm{i}(y)$, where
	  $\mathrm{i}(x)$ is a running index for nodes.
	  $\ell_E$ and $\ell_V$ are respectively the edge and node labeling functions. $p(x)$ is the set of node $x$'s properties, and $a(x,y)$ is the set of edge $(x,y)$'s attributes.
	}
\end{figure*}

\begin{figure*}[ht]
\centering
\includegraphics[width=.9\textwidth]{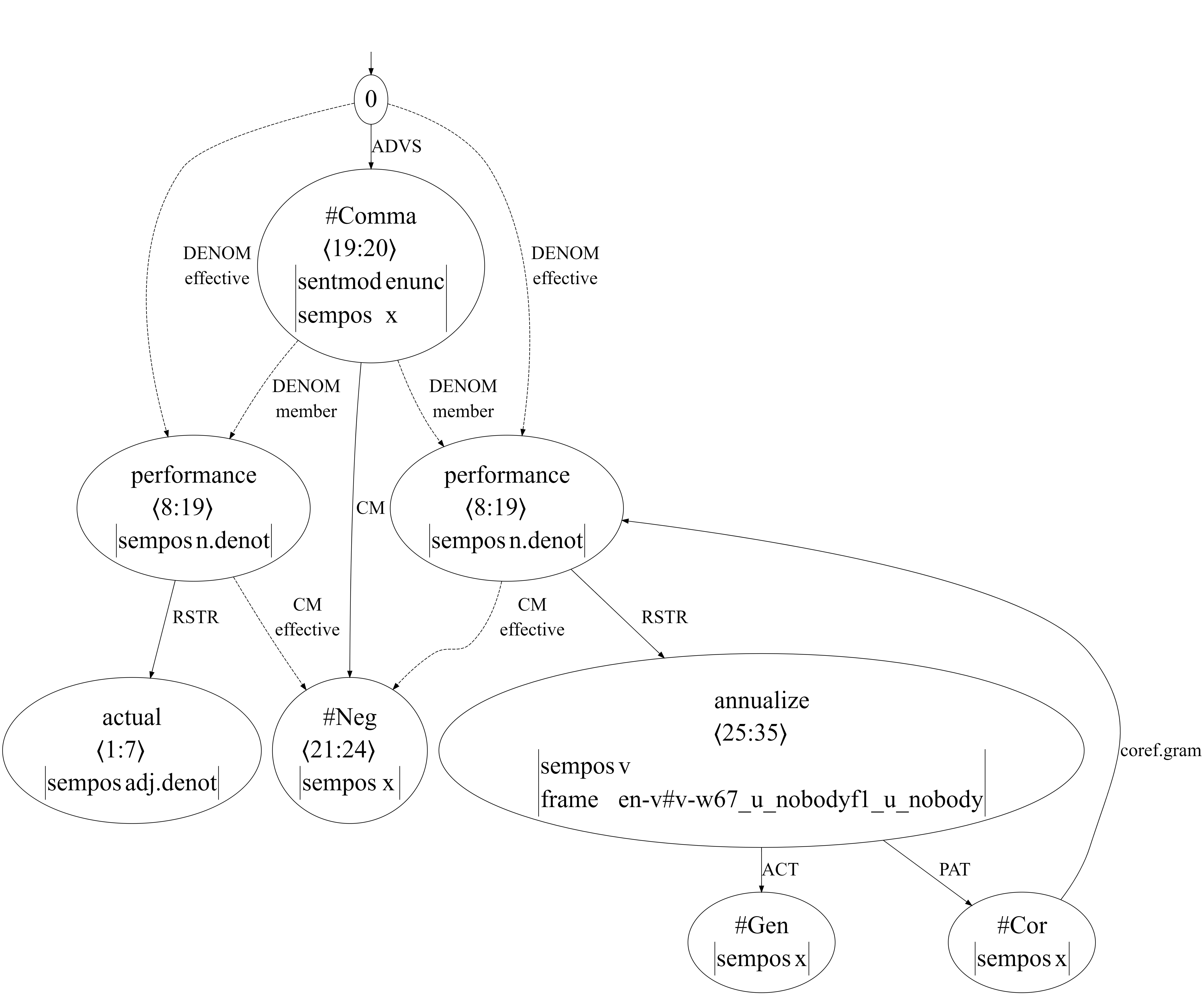}
  \caption{PTG graph, in the MRP formalism, for the sentence
  ``*Actual performance, not annualized''.
  Edge labels are shown on the edges.
  Node labels are shown inside the nodes, along with any node properties
  (in the form \texttt{|property value|}).
  Anchoring is also provided for PTG.}
  \label{fig:example_ptg}
\end{figure*}
\begin{figure*}[ht]
\centering
 \includegraphics[width=.9\textwidth]{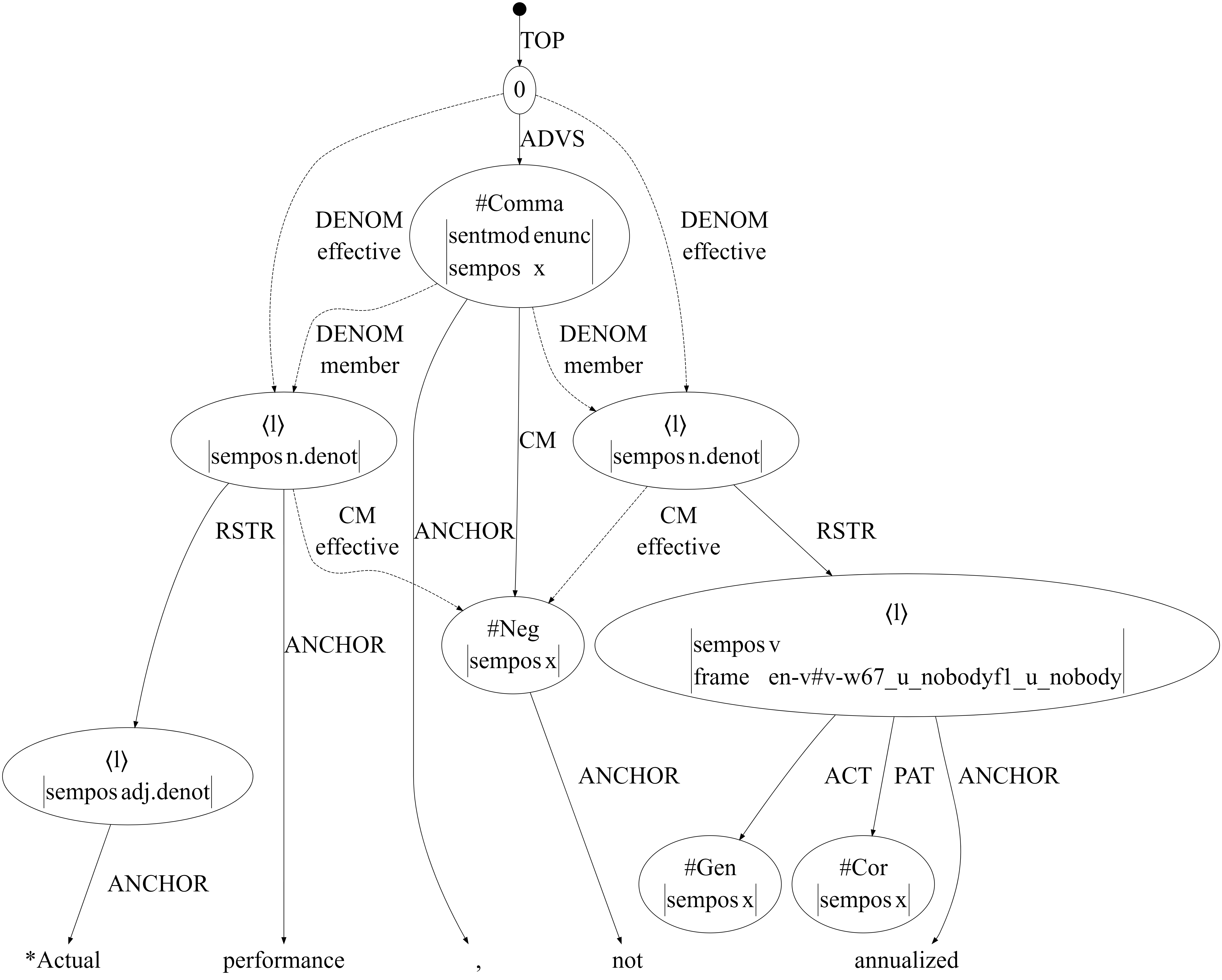}
  \caption{Converted PTG graph in the TUPA intermediate graph representation.
  Same as in the intermediate graph representation for all frameworks,
  it contains a virtual root node attached to the graph's top node with a \textsc{Top}
  edge, and virtual terminal nodes corresponding to text tokens,
  attached according to the anchoring
  with \textsc{Anchor} edges.
  Same as for all frameworks with node labels and properties (i.e., all but UCCA),
  labels and properties are replaced with placeholders corresponding to anchored tokens,
  where possible.
  The placeholder $\langle \ell \rangle$
  corresponds to the concatenated lemmas of anchored tokens.
  For graphs containing cycles, like this one, the cycles are broken by removing an arbitrary edge in the cycle (the \texttt{coref.gram} edge in this case).}
  \label{fig:example_ptg_tupa}
\end{figure*}

\section{TUPA-MRP}\label{sec:tupa}
TUPA \cite{hershcovich2017a} is a transition-based parser supporting general DAG parsing. The parser state is composed of a buffer $B$ of tokens and nodes to be processed, a stack $S$ of nodes currently being processed, and an incrementally constructed graph $G$. The input to the parser is a sequence of tokens: $w_1, \ldots, w_n$. A classifier is trained using an oracle to select the next transition based on features encoding the parser's current state, where the training objective is to maximize the sum of log-likelihoods of all gold transitions at each step.

The MRP variant \cite{hershcovich-arviv-2019-tupa}
supports node and edge labels, as well as node properties
and edge attributes. The code is publicly available.\footnote{\url{https://github.com/danielhers/tupa/tree/mrp}}

\subsection{Transition set}\label{sec:tupa-transitions}

The TUPA-MRP transition set, shown in Figure~\ref{fig:transitions}, is the same as the one
used by \newcite{hershcovich-arviv-2019-tupa}.
It includes the transitions \textsc{Shift} and \textsc{Reduce} to manipulate the stack, \textsc{Node$_X$} to create nodes compositionally, \textsc{Child$_X$} to create unanchored children, \textsc{Label$_X$} to label nodes, \textsc{Property$_X$} to set node properties, \textsc{Left-Edge$_X$} and \textsc{Right-Edge$_X$} to create edges, \textsc{Attribute$_X$} to set edge attributes, \textsc{Swap} to allow non-planar graphs and \textsc{Finish} to terminate the sequence.

\subsection{Transition Classifier}\label{sec:tupa-classifier}

To predict the next transition at each step, TUPA uses a BiLSTM module followed by an MLP and a softmax layer for classification \cite{kiperwasser2016simple}. The BiLSTM module is applied before the transition sequence starts, running over the input tokenized sequence. It consists of a pre-BiLSTM MLP with feature embeddings (\S\ref{sec:tupa-features}) and pre-trained contextualized BERT \cite{devlin-etal-2019-bert} embeddings concatenated as inputs,
followed by (multiple layers of) a bidirectional recurrent neural network \cite{schuster1997bidirectional,graves2008supervised} with a long short-term memory cell \cite{hochreiter1997long}.

Whenever a \textsc{Label$_X$}/\textsc{Property$_X$}/\textsc{Attribute$_X$} transition is selected, an additional classifier is evoked with the set of possible label/property/attribute values for the currently parsed framework, respectively, as possible outputs. This hard separation is made due to the large number of node labels and properties in the MRP frameworks.

\subsection{Features}\label{sec:tupa-features}

In both training and testing, we use vector embeddings representing the lemmas, coarse POS tags (UPOS) and fine-grained POS tags (XPOS). These feature values are provided by UDPipe as companion data by the task organizers. In addition, we use punctuation and gap type features \cite{maier-lichte:2016:DiscoNLP}, and previously predicted node and edge labels, node properties, edge attributes and parser actions. These embeddings are initialized randomly \cite{glorot2010understanding}.

To the feature embeddings, we concatenate numeric features representing the node height, number of parents and children, and the ratio between the number of terminals to total number of nodes in the graph $G$ \cite{hershcovich2017a}. Numeric features are taken as they are, whereas categorical features are mapped to real-valued embedding vectors. For each non-terminal node, we select a \textit{head terminal} for feature extraction, by traversing down the graph, selecting the first outgoing edge each time according to alphabetical order of labels.

\subsection{Intermediate Graph Representation}\label{sec:tupa-format}

We mostly reuse \newcite{hershcovich-arviv-2019-tupa}'s
internal representation of MRP graphs in TUPA,
where top nodes and anchoring are combined into the graph by adding a virtual root node
and virtual terminal nodes, respectively, during preprocessing.
Similarly, we introduce placeholders in the node labels and properties matching the tokens they are aligned to,
and collapse AMR name properties.
In the case of DRG and PTG, the newly added frameworks, where graphs may contains cycles, we break those cycles in order for them to be parseable by TUPA, which supports general DAG parsing. Only 0.27\% of the DRG graphs in the provided dataset are cyclic. In the case of PTG, 33.97\% are cyclic.
Figure~\ref{fig:example_ptg} shows an example PTG graph, and Figure~\ref{fig:example_ptg_tupa} the graph in TUPA's intermediate representation.
As the latter demonstrates, cycles are broken by removing an arbitrary edge in the cycle (the \texttt{coref.gram} edge in this case).


\subsection{Constraints}\label{sec:tupa-constraints}

As each framework has different constraints on the allowed graph structures,
we apply these constraints separately for each one.
During training and parsing, the relevant constraint set rules out some of the transitions
according to the parser state.

Some constraints are task-specific, others are generic.
For the new frameworks, DRG and PTG, all the constraints, except for one (PTG being multigraph), are derived from the graph properties as defined by their component pieces.\footnote{\url{http://mrp.nlpl.eu/2020/index.php?page=15}} For example, both require node labels, but only PTG requires node properties. No new types of constraints were needed to be added to TUPA to support these frameworks.

\begin{table*}[t]\centering\setlength{\tabcolsep}{3pt}
\begin{tabular}{lllrrcccl}
\multicolumn{1}{l}{\textbf{Track}} &
  \multicolumn{1}{l}{\textbf{Framework}} &
  \multicolumn{1}{l}{\textbf{System}} &
  \multicolumn{1}{l}{\textbf{\# Epochs}} &
  \multicolumn{1}{l}{\textbf{Best Epoch}} &
  \multicolumn{1}{l}{\textbf{Validation F1}} &
  \multicolumn{1}{l}{\textbf{Eval F1}} &
  \multicolumn{1}{l}{\textbf{Rank}} &
  \textbf{Best System} \\ \hline
CF & EDS     & HIT-SCIR & 6    & 2   & 0.82 & 0.80 & 5 & 0.94 (H) \\
CF & PTG     & TUPA     & 32   & 19  & 0.53 & 0.54 & 4 & 0.89 (H) \\
CF & UCCA    & TUPA     & 99   & 66  & 0.79 & 0.73 & 4 & 0.76 (Ú)    \\
\color{gray} CF & \color{gray}  UCCA    & \color{gray} HIT-SCIR      & \color{gray} 6   & \color{gray} 3  & \color{gray} 0.78 &&&  \\
CF & AMR     & TUPA     & 8    & 2   & 0.44 & 0.52 & 5 & 0.82 (H) \\
CF & DRG     & TUPA     & 200  & 99  & 0.52 & 0.63 & 5 & 0.94 (Ú)    \\
CL & PTG     & TUPA     & 20   & 13  & 0.60 & 0.58 & 4 & 0.91 (Ú)    \\
CL & UCCA    & HIT-SCIR & 13   & 6   & 0.77 & 0.75 & 4 & 0.81 (Ú)    \\
\color{gray} CL & \color{gray}  UCCA    & \color{gray} TUPA      & \color{gray} 100   & \color{gray} 95  & \color{gray} 0.43 &&&  \\
CL & AMR     & TUPA     & 21   & 12  & 0.44 & 0.45 & 4 & 0.80 (H) \\
CL & DRG     & TUPA     & 100 & (*) 68 & 0.52 & 0.62 & 4 & 0.93 (H) \\
\color{gray} CL & \color{gray} DRG     & \color{gray}TUPA     & \color{gray} 100 & \color{gray} 81 & \color{gray}0.51 &&&  \\ \hline
CF & Overall &          &      &     &       & 0.64 & 4 & 0.86 (H\&Ú)  \\
CL & Overall &          &      &     &       & 0.60 & 4 & 0.85 (H\&Ú)
\end{tabular}
\caption{Training details and official evaluation MRP F-scores.
For comparison, the highest score achieved for each framework and evaluation set is shown: H stands for Hitachi \cite{Oza:Mor:Kor:20} and Ú for ÚFAL \cite{Na:Min:20}. HIT-SCIR for English UCCA (CF) and TUPA for German UCCA (CL), both in gray, were not used in the submission, since their validation F1 were lower than the other system. For German DRG (CL) we trained 2 parsers: one on only the CL DRG dataset (in grey), not used in the submission, and another (*) trained on the English DRG dataset in per-training. The number of epochs does not include pre-training on English DRG.
\label{tab:official}}
\end{table*}

\subsection{Training details}\label{sec:tupa-training}

The model is implemented using DyNet v2.1
\cite{neubig2017dynet}.\footnote{\url{http://dynet.io}}
Unless otherwise noted, we use the default values provided by the package.
We use the same hyperparameters as \newcite{hershcovich-arviv-2019-tupa},
without any hyperparameter tuning on the CoNLL 2020 data.

We use the weighted sum of last four hidden layers of a BERT \cite{devlin-etal-2019-bert} pre-trained model\footnote{\url{https://github.com/huggingface/pytorch-transformers}} as extra input features,
summing over wordpiece vectors to get word representations.

\subsection{Cross-framework track}\label{sec:tupa-cf}
In the cross-framework track, we use the English \texttt{bert-large-cased} pre-trained encoder, and train separate TUPA models for each of the PTG, UCCA, AMR and DRG frameworks. Table~\ref{tab:official} shows the number of training epochs per framework, as well as validation and evaluation results.

\subsection{Cross-lingual track}\label{sec:tupa-cl}
For the cross-lingual track, as a generic contextualized encoder that supports many languages, we use multilingual BERT (\texttt{bert-base-multilingual-cased}) and train the models exactly the same as in the cross-framework track (separate model for each framework's respective monolingual dataset from the cross-lingual track), for Czech PTG and Chinese AMR.

For German DRG, as the provided dataset contains a relatively small amount of examples, 1575 as opposed to 6606 in English DRG (from the cross-framework track), we first pre-train a model on the DRG data in English and then fine-tune it on the DRG German dataset, in this case using mBERT to facilitate cross-lingual transfer. Surprisingly, this improves  our validation F1 score only by 0.013 points as opposed to training on the German dataset only, showing that the contribution of cross-lingual transfer is limited (but at least not detrimental) with this architecture and data sizes.

\section{HIT-SCIR Parser}\label{sec:hit-scir}
The HIT-SCIR parser \citep{che-etal-2019-hit} is a transition-based parser, which extended previous parsers by employing stack LSTM \cite{dyer2015transition} to allow computing homogeneous operation within a batch efficiently, and by adopting and fine-tuning BERT \citep{devlin-etal-2019-bert} embedding for effectively encoding contextual information.
The parser is implemented in the AllenNLP framework \cite{Gardner2017AllenNLP}.
It supports parsing DM, PSD, UCCA, EDS and AMR, all included in the 2019 MRP shared task. The official dataset would be pre-processed for system input and post-processed for output.

In our experiment, we modified the HIT-SCIR MRP 2019 parser to support the 2020 data for English EDS (for the cross-framework track) and German UCCA (for the cross-lingual track). We also explored the possibilities of employing multitask learning with the parser (\S\ref{sec:multitask}).  A repository containing our modified version of the parser is publicly available.\footnote{\url{https://github.com/ruixiangcui/hit-scir-mrp2020}}

\subsection{Transition set}\label{sec:hit-scir-transitions}

\newcite{che-etal-2019-hit} defined a different transition set per framework, according to framework's characteristics. 
As UCCA and EDS are already targets of 2019  MRP shared task, we inherit the existing transition sets for both frameworks. For UCCA, the transition system was modelled after that of the UCCA-specific (not MRP generic) TUPA \cite{hershcovich2017a}, which includes \textsc{Shift}, \textsc{Reduce}, \textsc{Node$_X$}, \textsc{Left-Edge$_X$}, \textsc{Right-Edge$_X$}, \textsc{Left-Remote$_X$}, \textsc{Right-Remote$_X$} and \textsc{Swap}.

The parser's EDS transition set is based on \citet{buys-blunsom-2017-robust}'s work, from which \textsc{Node-Start$_X$} and \textsc{Node-End} are two steps to create concept nodes and form node alignment. Apart from these two, \textsc{Shift}, \textsc{Reduce}, \textsc{Left-Edge$_X$}, \textsc{Right-edge$_X$}, \textsc{Drop}, \textsc{Pass} and \textsc{Finish} are also used to represent EDS transition process.

\subsection{Transition Classifier}
The parser state is represented by $(S,L,B,E,V)$, where $S$ is a
stack holding processed words, $L$ is a list holding words popped out of $S$ that will be pushed back in the future, and $B$ is a buffer holding unprocessed words. $E$ is a set of labeled dependency arcs. $V$ is a set of graph nodes include  concept  nodes and surface tokens. Transition classifier takes $S, L, B$ and also the action history as input, all are modeled with stack LSTM, and outputs an action. The input to the parser is a sequence of BERT embedding. A transition classifier takes $S,L,B$ and the action history as inputs and maximizes the log-likeihood of the correct action given the current state using an oracle to get the correct action.

\subsection{Preprocessing}\label{sec:hit-scir-preprocessing}
MRP 2019 provided companion data (containing the results of syntactic preprocessing) in both \texttt{CoNLL-U} and \texttt{mrp} formats. However, this year's task only provides \texttt{mrp}-formatted companion data.
Since the HIT-SCIR 2019 parser can only take \texttt{CoNLL-U}-formatted companion data, we update it to allow converting companion data provided by 2020 MRP shared task from \texttt{mrp} format to \texttt{CoNLL-U} format.

\subsection{Anchoring}\label{sec:hit-scir-anchoring}
The parser itself is also modified to support the MRP 2020 task. For EDS parsing specifically, in this year's task's provided data, anchoring for a token containing spaces, such as an integer number followed by a fraction number (e.g., ``3 1/2'') is treated as one token, while the original parser's node anchoring treats the two parts separately.
Another example would be: ``x-Year-to-date 1988 figure includes Volkswagen domestic-production through July.'' In this sentence, ``x-Year-to-date 1988'' is marked as a node anchored from characters 2 to 26, but the provided companion data treats ``x-Year-to-date'' as anchored from characters 0 to 14 as the corresponding token anchor. To handle these cases, we allow the parsing system to perform partial node alignment regardless of overlapping token anchors.

\subsection{Constraints}\label{sec:hit-scir-constraints}
The second problem we encounter when parsing EDS is that there are a few instances that are too short, and no valid actions can be performed according to the existing transition system. In this case, we allow the \textsc{Finish} action, adding it directly to the allowed action set when no valid action exists, with the effect that the transition sequence is terminated and the current graph is returned.

\subsection{Training}\label{sec:hit-scir-training}
We train the modified HIT-SCIR parser on English and German UCCA (in the cross-framework and cross-lingual tracks, respectively) and English EDS (in the cross-lingual track).
The training time is 2 days 1 hour for English UCCA, 22 hours for German UCCA, and 4 days 6 hours for English EDS.
The training details are shown in Table~\ref{tab:official}. Since HIT-SCIR parser's validation score on cross-framework UCCA is 0.01 lower than TUPA, we opt for TUPA in that category.
Hyperparameters are taken directly from \newcite{che-etal-2019-hit}.

\section{Results}\label{sec:results}

Table~\ref{tab:official} presents the
averaged scores on the test sets in the official evaluation,
for our submission and for the best-performing system in each framework and evaluation
set. 

\paragraph{Validation vs. evaluation scores.}

The validation scores of 5 out of the 9 parsers is lower than their evaluation score: CF PTG by 0.01 F1 points, CF AMR by 0.08, CF DRG by 0.11, CL AMR by 0.01 and CL DRG by 0.1. We hypothesize it is due to the randomness in the evaluation metric: the MRP scorer uses a search algorithm to find a correspondence relation between the gold-standard and system graphs that maximizes tuple overlap. This search algorithm runs for a limited number of iterations. In order to decrease its running time, we used a lower limit on its parameters (10 random restarts, 5,000 iterations) than the default (20 random restarts, 50,000 iterations), which may have affected the accuracy of our validation score and potentially our system performance.

\paragraph{CF vs. CL tracks.}

Surprisingly, the CL track scores are mostly on-par with the CF tack ones, even though the CL parsers were often trained on significantly less examples. While the CF UCCA training dataset contains 6,872 examples and the CL UCCA contains only 3,713, both parsers gained similar scores. Similarly, the CF DRG dataset contains 6,606 example, while the the CL DRG contains only 1,575. TUPA trained only on the 1,575 examples gained a similar score to the CF one, while training on less then a fourth of the examples.
The CF PTG dataset contains 42,024 examples. And while the CL PTG contains a lower, however similar, amount (39,560), it got a higher score (0.07 F1 point in validation, and 0.04 in evaluation). And while the CL AMR dataset is only a third of the CF AMR datsaet (16,529 and 57,885 examples respectively), both parser gained the same validation score. However, the evaluation score of the CF AMR is higher by 0.07 F1 points. This could be possibly attributed to our MRP scorer low iteration limit.



\begin{table}[t]
\centering
\begin{tabular}{lr}
\multicolumn{1}{l}{\textbf{Hyperparameter}}                                     & \multicolumn{1}{l}{\textbf{Value}} \\ \hline
Task embedding dim                                                                & 20                                  \\ \hline
\textit{Shared encoder} \\
Input dim               & 1024                                \\ \hline
\textit{Framework-specific encoder} \\
Input dim            & 768                                 \\ \hline
\textit{Both encoders} \\
Input dim               & 768 \\
Projection dim         & 512 \\
Feedforward hidden dim  & 512 \\
\# layers & 3                                   \\
\# attention heads & 8  
\end{tabular}
\caption{HIT-SCIR multitask model hyperparameters.}
\label{tab:Multitask}
\end{table}

\begin{table*}[th]
\centering
\begin{tabular}{lrrccc}
&&&\textbf{Validation}&\textbf{Validation}&\textbf{Validation}\\
\textbf{Sharing architecture} &
  \textbf{\# Epochs} &
  \textbf{Best Epoch} &
  \textbf{Average F1} &
  \textbf{UCCA F1} &
  \textbf{EDS F1} \\ \hline
Shared encoder \\ + task embedding &
  13 &
  2 &
  0.55 &
  0.68 &
  0.43 \\ + task specific encoders &
  13 &
  4 &
  0.38 &
  0.49 &
  0.27
\end{tabular}
\caption{HIT-SCIR multitask model training details and scores.}
\label{tab:Multitask_score}
\end{table*}

\begin{figure*}[th]
   \centering\includegraphics[width=.9\textwidth]{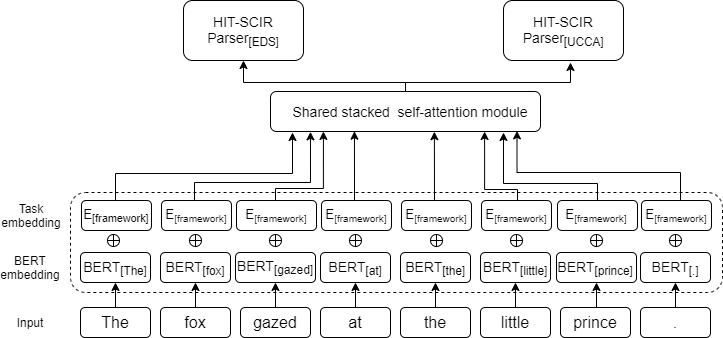}
\caption{Illustration of the first variant of the HIT-SCIR multitask model,
parsing the sentence
``The fox gazed at the little prince.''
Top: Dedicated HIT-SCIR parsers for each framework.
		Bottom: Encoder architecture.
		BERT embeddings are extracted for each token and are concatenated with framework-specific learned embedding.
		Vector representations for the input tokens are then computed
		by a shared stacked self-attention encoder.
		The encoded vectors are then fed to a framework-specific HIT-SCIR parsers as input tokens.
}\label{fig:hit_scir_multitask_model_1}
\end{figure*}

\begin{figure*}[th]
   \centering\includegraphics[width=.9\textwidth]{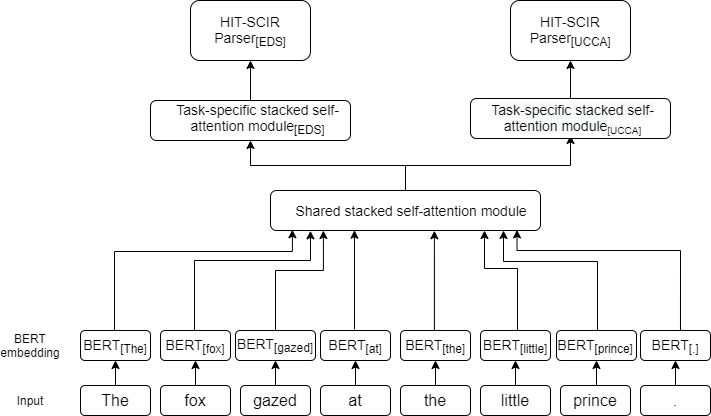}
\caption{Illustration of the second variant of the HIT-SCIR multitask model,
parsing the sentence
``The fox gazed at the little prince.''
Top: Dedicated HIT-SCIR parsers for each framework.
		Bottom:  Encoder architecture.
		BERT embeddings are extracted for each token. Vector representations for the input tokens are then computed by a shared stacked self-attention encoder and by a framework-specific self-attention encoder.
		The encoded vectors are then fed to a framework-specific HIT-SCIR parsers as input tokens.
}\label{fig:hit_scir_multitask_model_2}
\end{figure*}

\section{Multitask Cross-Framework Parsing}\label{sec:multitask}
In addition to training separate models per framework and language, we also experiment with training multitask cross-framework parsers, using a neural architecture with parameter sharing \cite{peng-etal-2017-deep,Peng-EtAl:2018:NAACL,Her:Abe:Rap:18,lindemann-etal-2019-compositional,hershcovich-arviv-2019-tupa}.
We use the HIT-SCIR parser as a basis, with different variations of shared architecture on top of it. For our experiments we choose the UCCA and EDS frameworks. The code is publicly available.\footnote{\url{https://github.com/OfirArviv/hit-scir-mrp2020/tree/multitask}}

\subsection{Model}\label{sec:multitask-model}
We try two different sharing architectures.
In both architectures, both frameworks share a stacked self-attention encoder (see Table~\ref{tab:Multitask} for details). In the first variation, we additionally use task embeddings; in the second, we use task-specific encoders instead.

\paragraph{Task embedding.}
In the first sharing architecture, both frameworks share a stacked self-attention encoder whose input is a BERT embedding concatenated with a learned task embedding of dimension 20. This has been shown to help in shared architecture multitask models \cite{Sun2020ERNIE2A}, as well as cross-lingual parsing models, where a language embedding is used \cite{ammar-etal-2016-many,de-lhoneux-etal-2018-parameter}.
In our case, the ``task'' has two possible values, namely UCCA and EDS.
The output of the shared encoder is then fed into two separate ``decoders'', which are HIT-SCIR parser transition classifiers. We use one for each framework, whose architecture and hyperparameters are the same as in the single task setting.
Figure~\ref{fig:hit_scir_multitask_model_1} illustrates this architecture.

\paragraph{Task-specific encoders.}
In the second architectures, both frameworks share a stacked self-attention encoder whose input is a BERT embedding, and in addition each framework has another stacked self-attention encoder of it own,
similar in concept to \newcite{peng-etal-2017-deep,Peng-EtAl:2018:NAACL}'s \textsc{freda1} architecture (which, however, used BiLSTMs), also employed by \newcite{Her:Abe:Rap:18,lindemann-etal-2019-compositional}.
The outputs of these encoders are processed the same as in the first variation (task-specific decoders). Figure~\ref{fig:hit_scir_multitask_model_2} illustrates this architecture.

\subsection{Training details}\label{sec:multitask-training}

Each training batch contains examples from a single framework, while the model is alternating between the different batch types. As the EDS training dataset is much larger than the UCCA one, we balance them out by training the same number of examples from each framework in each epoch.
Due to time constraints we tried out only a single set of hyperparameters, chosen arbitrarily without tuning.
We select the epoch with the best average MRP F-score on a validation set, which is the union of both validation sets of EDS and UCCA.

\subsection{Results}\label{sec:multitask-results}
Table \ref{tab:Multitask_score} presents the average scores on the validation sets for multitask trained models. The multitask HIT-SCIR consistently falls behind the single-task one, for each framework separately and in the overall scores;
but it is clear that our first multitask architecture (with task embedding) outperforms the second one (with task-specific encoders). 

\subsection{Discussion}\label{sec:multitask-discussion}
Previous results on multitask MRP showed mixed results, some showing improved performances \cite{peng-etal-2017-deep,Her:Abe:Rap:18,lindemann-etal-2019-compositional}. Others failed to show improvements \cite{hershcovich-arviv-2019-tupa}, and argued that the large multitask models were underfitting due to insufficient training. In our case, however, the multitask models underperform despite reaching convergence.

We hypothesize that with better hyperparameters or different sharing architectures, more favorable results could be obtained.
However, it is possible that multitask learning would be more helpful in a factorization-based parser \cite{peng-etal-2017-deep,lindemann-etal-2019-compositional}, where inference is global and more uniform across frameworks.
A transition-based parser may be less suited for utilizing information from different tasks that have different transition systems, as in the HIT-SCIR parser. Adapting it to have a more uniform transition system, like TUPA does, could facilitate cross-framework sharing. Alternatively, improving TUPA's training efficiency would also enable such experimentation.

\section{Conclusion}\label{sec:conclusion}
We have presented TUPA-MRP and a modified HIT-SCIR parser, which constitute the HUJI-KU submission in the CoNLL 2020 shared task on Cross-Framework Meaning Representation. TUPA is a general transition-based DAG parser with a uniform transition system, which is easily adaptable for multiple frameworks. We used it for parsing in both the cross-framework and the cross-lingual tracks, adapting it for the newly introduced frameworks, PTG and DRG.  HIT-SCIR is a transition-based parser with framework-specific transition systems, which we adapted for this year's shared task and used for English EDS and UCCA parsing in the cross-framework track. The HIT-SCIR parser was additionally used in experimenting on multitask learning, with negative results for that approach.

Future work will tackle the MRP task with more modern transition-based-like parser architectures, such as pointer networks \cite{ma-etal-2018-stack}, which have so far only been applied to bilexical framworks, i.e., flavor-0 SDP \cite{fernandez-gonzalez-gomez-rodriguez-2020-transition}.

\section*{Acknowledgments}

We are grateful for the valuable feedback from the anonymous reviewers.

\bibliography{mrp,sdp,ltg,conll,references}
\bibliographystyle{acl_natbib}

\end{document}